\documentclass[10pt,twocolumn,letterpaper]{article}

\usepackage{cvpr}
\usepackage{times}
\usepackage{epsfig}
\usepackage{graphicx}
\usepackage{amsmath}
\usepackage{amssymb}

\usepackage{blindtext}
\usepackage{caption}

\usepackage[pagebackref=true,breaklinks=true,letterpaper=true,colorlinks,bookmarks=false]{hyperref}

\usepackage[table]{xcolor}
\usepackage{makecell}

\newcommand{\one}[0]{\includegraphics[height=.012\textwidth]{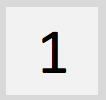}}
\newcommand{\two}[0]{\includegraphics[height=.012\textwidth]{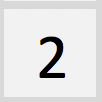}}
\newcommand{\three}[0]{\includegraphics[height=.012\textwidth]{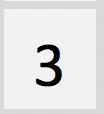}}
\newcommand{\four}[0]{\includegraphics[height=.012\textwidth]{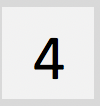}}
\newcommand{\five}[0]{\includegraphics[height=.012\textwidth]{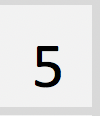}}
\newcommand{\six}[0]{\includegraphics[height=.012\textwidth]{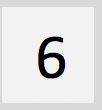}}
\newcommand{\seven}[0]{\includegraphics[height=.012\textwidth]{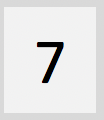}}
\newcommand{\eight}[0]{\includegraphics[height=.012\textwidth]{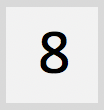}}
\newcommand{\nine}[0]{\includegraphics[height=.012\textwidth]{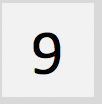}}
\newcommand{\ten}[0]{\includegraphics[height=.012\textwidth]{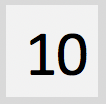}}
\newcommand{\eleven}[0]{\includegraphics[height=.012\textwidth]{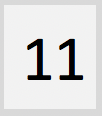}}

\cvprfinalcopy 

\def\cvprPaperID{4273} 

\ifcvprfinal\fi
\begin{document}
\linespread{0.95}
\title{Deep Multi-Modal Sets}

\author{Austin Reiter\\
Facebook AI Research\\
{\tt\small areiter@fb.com}
\and
Menglin Jia\\
Cornell University\\
{\tt\small mj943@cornell.edu}
\and
Pu Yang\\
Facebook AI Research\\
{\tt\small puyang07@fb.com}
\and
Ser-Nam Lim\\
Facebook AI Research\\
{\tt\small sernamlim@fb.com}
}

\maketitle

\begin{abstract}
   Many vision-related tasks benefit from reasoning over multiple modalities to leverage
   complementary views of data in an attempt to learn robust embedding spaces.  Most deep 
   learning-based methods rely on a late fusion technique whereby multiple feature types are encoded 
   and concatenated and then a multi layer perceptron (MLP) combines the fused embedding 
   to make predictions.  This has several limitations, such as an unnatural enforcement that all features be present at 
   all times as well as constraining only a constant number of occurrences of a feature modality at any given time.  Furthermore,
   as more modalities are added, the concatenated embedding grows.  To mitigate this, we propose
   \textbf{Deep Multi-Modal Sets}: a technique that represents a collection of features as an unordered 
   \textbf{set} rather than one long ever-growing fixed-size vector.  The set is constructed so that we have invariance both to
   permutations of the feature modalities as well as to the cardinality of the set.  We will also show that with particular choices in our model architecture, we can yield \textbf{interpretable feature performance} such that during inference time we can observe which modalities are most contributing to the prediction.  With this in mind, we demonstrate a scalable, multi-modal framework that reasons over
   different modalities to learn various types of tasks.  We demonstrate new state-of-the-art performance
   on two multi-modal datasets (Ads-Parallelity \cite{zhang2018equal} and MM-IMDb \cite{Arevalo17}).
\end{abstract}

\section{Introduction}

Traditional vision tasks typically formulate problems with a single input (e.g., an image) to infer a desired output (e.g., classifying a scene, detecting objects, etc).  More recently, the advantages of multiple complementary inputs to achieve a desired output has become more popular, especially for higher-level reasoning tasks such as Visual Question Answering (VQA) and video-based tasks.  These types of models are referred to as \textit{multi-modal} models.


In multi-modal models, the goal is to construct a model that is able to leverage different types of information with a common goal in mind.  A very typical manifestation of this is to combine both visual and textual information.  Here, the model seeks a way to \textit{fuse} these data sources, usually by leveraging their individual discrimination capabilities, and then combining them together into a single representation for the final prediction task.  As such, an image model (e.g., CNN) may learn to represent raw images as feature embeddings while a text model (e.g., Text-CNN, LSTM, etc) is similarly learned to represent raw text, and these two are ``fused'' (or concatenated) into a single feature.  This fused feature is used for any final down-stream tasks.

\begin{figure*}[t]
    \centering
    \includegraphics[width=1\textwidth]{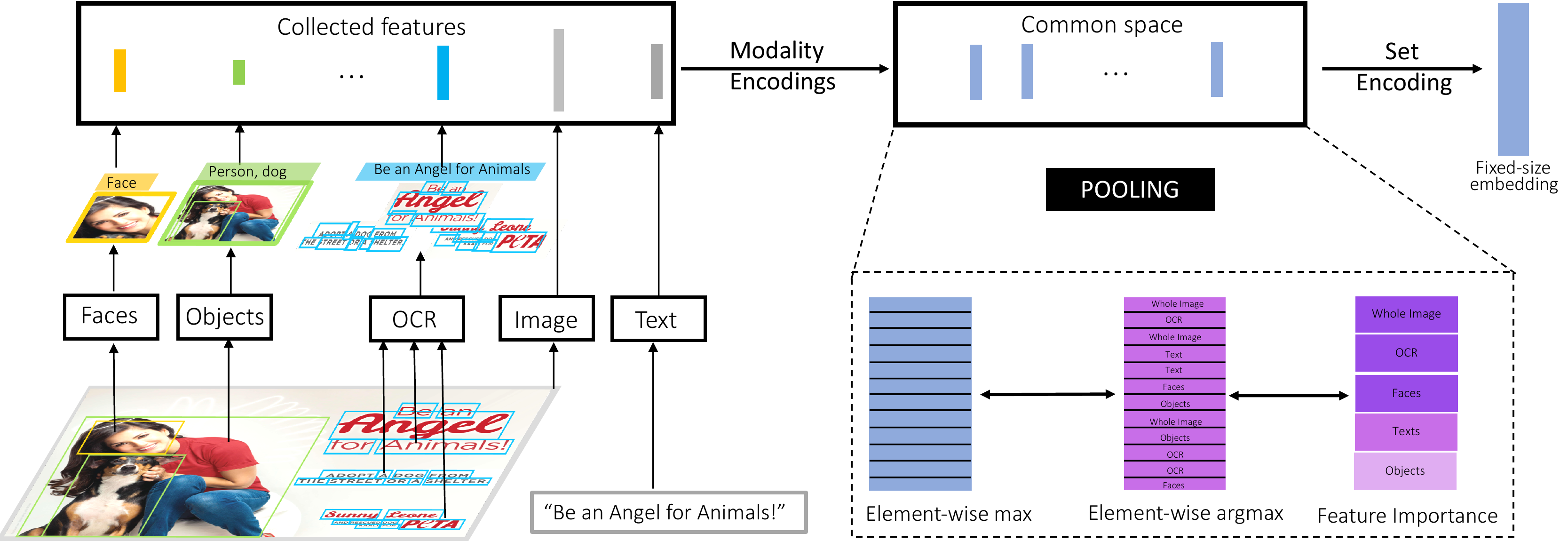}
    \captionof{figure}{\textbf{Overview of Deep Multi-Modal Sets.}  An image (bottom-left), possibly along with additional raw inputs (e.g., raw text; ``Be an Angel for Animals!''), is/are processed through various types of detectors, such as face and object detectors as well as OCR extraction.  Additionally, we encode raw inputs such as the entire image through a CNN and raw text through an NLP encoding model.  Each is collected and re-encoded to a common space and then pooled to a fixed-size embedding.  We call this the ``Set Encoding''.  When we use an interpretable feature pooling scheme, such as \textit{max-pooling}, we obtain a metric of feature importance (see more in text).}
    \label{fig:teaser}
\end{figure*}

With this model architecture in mind, we sought out to explore the following question: \textbf{what if we approached visual modeling as a multi-modal problem?}  More specifically, rather than encoding all visual information from a photo into a single, all-encompassing feature embedding (as is typical, for example, as we do with ImageNet-like tasks), instead we look to break an image down into more \textit{compositional components}, such as objects, faces, overlaid text, etc, and then learn individual representations for each to be fused into a multi-modal framework.  Can we do better like this?  In order to explore an answer to this question, we first must realize that a standard concatenation model is too limiting to support features such as objects or faces, where there may be 0, 1, or variable number of occurrences on any given frame, and this does not stay constant across samples.  Therefore, a new way of thinking about how to fuse various modalities is required. Further, any modalities besides the image components mentioned, such as accompanying text and even audio, should also be easily added to the final embeddings.

To answer this question, this paper offers the following 3 primary contributions:

\begin{itemize}
    \item A new multi-modal architecture which utilizes \textit{unordered sets}; this adds much more flexibility in terms of number of feature inputs as well as number of feature occurrences per-sample.
    \item This new architecture allows us to start to think about compositional reasoning via multi-modal modeling
    \item An interpretable feature importance metric, allowing us to inspect our model at inference time for which feature modalities are most contributing to a given prediction.
\end{itemize}

The remainder of the paper is laid out as follows: first we review prior work on multi-modal modeling within the vision and machine learning literatures.  Then we provide an overview of our method, called \textit{Deep Multi-Modal Sets}, and conclude with experimental results and conclusions.

\section{Related Work}

Early and late fusion methods for multi-modal models have been explored for several years \cite{Atrey10}, for example, by combining low-level features with prediction-level features.  It has been shown in some scenarios that late fusion methods outperform early fusion methods, but not in all cases \cite{Snoek05}.  ``Late fusion'' is commonly defined as a combination of the output scores (or embeddings) from each unimodal branch of a multi-modal model.  Previous to deep learning, methods for combining these outputs ranged from weighted averaging \cite{Natarajan12} to rank minimization \cite{Ye12}.

Combinations of early and late fusion methods were proposed in \cite{Yang16}, whereby low-level and high-level features were fused via boosting.  Attention models have become recently popular, rooted with early work from \cite{Jacobs91} showing how a model is able to pick from a selection of networks via gating, based on the input.  Multi-modal attention has also been extended to temporal problems, as shown in \cite{Hori17,Long18}.

As much of the field has moved towards deep learning, given superior results in all domains from computer vision to natural language processing, it makes sense to focus in on methods built on deep models.  \cite{prezra2019mfas} proposed a method to search for an optimal neural architecture that optimizes fusing feature modalities based on an optimization scheme that poses the neural architecture search problem as a combinatorial search.

The work in \cite{kiela2018efficient} is particularly interesting and relevant to the current proposed work.  In this paper, the authors investigate various methods to combine text, as a discrete representation, along with a vision-based, more continuous modality, in a multi-modal framework.  The target of this work was to develop methods that are appropriate for large quantities of data that must be processed with light-weight, computationally-efficient operations.  In doing so, the authors touched upon fusion approaches that are very similar to unordered sets.  We note the parallels to this work and ours, where we are further generalizing and pushing these ideas to a more formal framework for multi-modal fusion.  As a follow-up to this work, \cite{kiela2019supervised} introduces a multi-modal bitransformer model which seeks to highlight the strengths of the text signal while supplementing with CNN-type image-only features, showing significant SOTA performance on difficult multi-modal benchmarks.

In \cite{zhang2018equal}, the question is explored as to whether two modalities, images and text, are complementary to one another and how to better understand their relationship.  An ensemble of SVM predictors are proposed to exploit the relationship between an image and associated text, all with a goal towards capturing ``parallelity''.  Multiple modalities modeled from deep neural networks were explored in \cite{Ngiam11} and \cite{Srivastava12}. Related to this, gated mechanisms and compact bilinear pooling were proposed as alternative ways to fuse modalities in \cite{Arevalo17, Fukui16}. 

Multi-modal models have also recently been proposed as a way to provide explanations for model decisions.  In \cite{Park18}, two datasets and a new model are proposed to provide joint textual rationales with attentional justifications for model decisions in tasks that ask questions about content in photos.  Other similar work \cite{Yang16,Xu16} has similarly focused on attention models in a multi-modal fashion for VQA tasks.

\section{Method Overview}

We begin with an overview of our Deep Multi-Modal Sets methodology.  First, we describe the generic \textit{multi-modal} problem as well as the baseline approach to this.  Next, we describe the Deep Sets overview, and conclude with Deep Multi-Modal Sets. 

\subsection{The Multi-Modal Problem}
\label{sec:multi_modal}


A multi-modal problem is one in which more than one feature type, or \textit{modality}, is given.  The goal is to combine these features together within a single model towards a common task.  More formally, consider that we have modalities $i \in \mathbf{I}$, represented by feature embeddings $X_i$.  In general, each $X_i$ may be of dimensions $(N_i \times M_i)$, for $N_i \geq 0$ occurrences of modality $i$, each of which is ($1 \times M_i$) in it's embedding space.  In this way, a feature modality may occur 0, 1, or more times in a given data sample.  

In the most simple form of a multi-modal model, a \textit{concatenated} feature is formed by appending all features into a single vector: $X_C = concat([X_1, \dots X_I])$, where $concat$ is a function that appends all 1-D vectors to one-another.  In this way, $X_C \in$ $\mathbb{R}^M$, where $\mathbf{M} = \sum_{i=1}^{I} N_i*M_i$.  We then typically attach an MLP to the end of $X_C$ to predict the target task.  There are two complications that may arise with this model in mind: 1) if $N_i = 0$ for any $i \in \mathbf{I}$; and  2) if $N_i > 1$ for any $i \in \mathbf{I}$.  For 1) it is typical to use a \textit{placeholder}, such as zeros, to indicate that the feature is missing; however, this can be unnatural to force into the model.  As for 2), there are no wide-spread methods to deal with multiple occurrences of a given feature modality, unless it's constant throughout.  However, if this varies across data samples, it is unknown how best to deal with this (perhaps you pad with the maximum number of occurrences, which is quite wasteful).  We will describe our technique to address these issues in the following sub-sections.

Another common issue with this formulation for multi-modal modeling is when there is a large imbalance in feature dimensions amongst the modalities.  When this happens, it is possible those feature types with higher dimensions dominate the model over those with less.  This can be addressed by encoding each modality individually before concatenating.  For the purposes of this work, we will denote $\phi_i(X_i)$ as an encoder acting on modality $i$.  The goal here is then to encode each modality to a common dimensionality $D$: $\phi_i(X_i): \mathbb{R}^{M_i} \to \mathbb{R}^D$ for all $i \in \mathbf{I}$.

Finally, if we consider the case where $\textbf{M}$ is very large, either due to many different modalities or many different occurrences of individual modalities, the final multi-modal model can get quite large due to the fact that an MLP is fully-connected.  To counter this, we would instead prefer an architecture that scales better with the cardinality of $\mathbf{I}$ (denoted as $\mathbf{|I|}$) and $\textbf{M}$.  Towards this end, we describe a new class of models that we may adopt to the multi-modal domain.

\subsection{Deep Sets}
\label{sec:deep_sets}

Deep Sets \cite{Zaheer17} refer to a class of models for machine learning tasks that are defined on \textit{sets}, in contrast to traditional approaches that operate on fixed dimensional vectors.  The main contribution of \cite{Zaheer17} is to define an architecture with properties that guarantee invariances across permutations of a collection of features as well as the cardinality (e.g., number of elements) within them; this collection is referred to as a \textit{set}.

Consider, as input to such a model, a set of vectors $X = \{x_1, \dots, x_S\}$.  We construct a function $\psi(X)$ that is ``indifferent'' to the ordering as well as the count of the elements in $X$.  Such a function would be able to map any number of elements to a fixed dimensional representation for use in a down-stream modeling task.  One example is the well-known \textit{sum-pooling} operator:

\begin{equation}
    \psi_{sum}(X) = \sum_{x \in X} \phi(x)
\end{equation}

\noindent where $\phi(x)$ is a transformation on input $x$ to an embedded space. 

In general, any operator applied to $X$ that yields an identical solution for any permutation $\pi: \psi(\{x_1, \dots, x_S\}) = \psi(\{x_{\pi(1)}, \dots, x_{\pi(S)}\}$ is sufficient for this model.  Another example is the \textit{max-pooling} operator:

\begin{equation}
    \psi_{max}(X) = \max_{x \in X} \phi(x)
    \label{eqn:max_pool}
\end{equation}

\noindent Here, the $\mathbf{max}$ operator is element-wise, in the sense that a set of $S$ elements, each of which is $(1 \times D)$, when operated on by the function in Eq. \ref{eqn:max_pool}, yields a $(1 \times D)$ vector at all times.  In the end, the output of the pooling operation is input to a down-stream model, such as an MLP, to classify label $y$ from input $X$ as:

\begin{equation}
    y = \rho(\psi(X))
\end{equation}

\subsection{Deep Multi-Modal Sets}

The primary contribution of this work is to incorporate the concepts of Deep Sets into the multi-modal modeling problem.  All of the shortcomings of the baseline model methodologies described in Sec.~\ref{sec:multi_modal} can be addressed by applying the concepts described in Sec.~\ref{sec:deep_sets}, with a few additional advantages which we will describe later on.  We call these \textit{Deep Multi-Modal Sets} (see Fig.~\ref{fig:teaser}).

The idea of concatenating feature modalities to one another can be thought of as a form of a ``pooling operator'', except that they do not conform to order invariance and it is very sensitive to number of modalities.  However, at its core, concatenation serves the purpose of combining several modalities into a single representation in order to perform down-stream modeling tasks.  With this in mind, we propose replacing the concatenation operation with a Deep Set; we collect all feature modalities into a single \textit{set}, and we pool them in order to extract a single embedding that jointly represents all features.  In this way, this model has the following advantages:

\begin{itemize}
    \item No matter how many modalities we have, or occurrences of individual modalities, the size of the model stays constant.  This scales very nicely with increasing number of features and modalities.
    \item If a particular feature is not present for a given frame, we do not force a place-holder; we simply do not include it in the set.
\end{itemize}

Using the notation from Sec.~\ref{sec:multi_modal}, given our feature modalities, we construct set $\mathbf{X} = \{X_i, i \in \mathbf{I}\}$.  Given that many feature modalities may naturally have different dimensionalities, we enforce encoders  $\phi_i(X_i)$ for each $X_i$ so that every feature occurrence in $\mathbf{X}$ has dimensionality ($1 \times D$).  For any modality with multiple occurrences, each occurrence becomes an individual element in $\mathbf{X}$.  In this way, the number of elements in $\mathbf{X}$ is equal to $\sum_{i \in I} N_i$.

The goal of the multi-modal model is to then learn a pooling operator $\psi(\mathbf{X})$.  In the end, we attach an MLP to the output of this operator, denoted as $\rho(\psi(X))$, to perform the final down-stream target task.  

\subsubsection{Feature Importance}
There are several choices for the pooling operator that one can use.  We listed two above in the sum and max pooling approaches.  Other popular choices are mean, min, and median pooling.  However, there are side effects to certain pooling operators.  For example, if the max pooling operator is used, we can get a sense of \textit{feature importance} in a way that is interpretable during inference time.  We show this depiction in Fig.~\ref{fig:teaser}.  In short, max pooling operates over an $(N \times d)$ tensor to produce a $(1 \times d)$ vector for any $N$ by extracting the maximum element for each of the $d$ dimensions, across all $N$ samples.

For the proposed model, this can be useful because each of the $N$ samples in the tensor (over which we are pooling) is an individual instance of a specific, known feature modality.  Therefore, as each max element is extracted, we keep track (per-dimension) of \textit{which} modality is contributing to this max-pooled feature (the so-called \textit{argmax}).  Those modalities which occur the most for a given forward pass during inference-time are deemed the most important for that set.  We can use this analysis to observe which features are more or less important for a given modeling problem.

The use of highly activated intermediate features as an importance measure has been shown in the vision literature \cite{ZhouCVPR15}.  In short, when we max-pool, the model is selecting features amongst the modalities for down-stream predictions via $\rho$.  We say the chosen features are ``important'' because they are ``surviving'' the pooling process; max-pooling is selecting features rather than combining (for example, as sum-pooling does), and these are the features that are used down-stream amongst the other features in the set.  In various works such as \cite{ZhouCVPR15}, it is shown that this sort of paradigm is highlighting \textit{attention} to important parts of the inputs that are most significantly impacting the predictive outcomes.

\section{Experiments and Discussions}
The proposed Deep Multi-Modal Sets are evaluated on two multi-modality datasets, namely MM-IMDb~\cite{Arevalo17} and Ads-Parallelity Dataset~\cite{zhang2018equal}. 

\subsection{Datasets}
\textbf{Ads-Parallelity Dataset.}
This dataset contains images and slogans from persuasive advertisements, for understanding the implicit relationship (parallel and non-parallel) between these two modalities.
This binary classification task is to predict if the image and text in the same advertisement convey the same message. 
A total of 327 samples are used (after removing duplicated images).
We use 5-fold cross-validation, and report both overall and per-class average accuracy following~\cite{zhang2018equal}.
Additionally we also report ROC AUC score.

\textbf{MM-IMDb.} Multimodal IMDb (MM-IMDb) dataset~\cite{Arevalo17} contains 25,959 movies including their plots, posters, and other metadata.
The task is to predict genres for each movie.
Since one movie can have multiple genres, this task is a multi-label classification problem.
We use the original split from the dataset, and report the F1 scores (micro, macro, and samples) on the given test set.


\subsection{Features}

\begin{table}[t]
\begin{center}
\resizebox{1\columnwidth}{!}{%
\begin{tabular}{ lr }
\Xhline{2\arrayrulewidth}
\multicolumn{1}{l}{\textbf{Features}} & \multicolumn{1}{|c}{\textbf{Inputs types}}\\
\Xhline{2\arrayrulewidth}
\multicolumn{1}{l}{WSL} & \multicolumn{1}{|c}{image (i), detected bounding boxes (bbox)}\\
\hline
\multicolumn{1}{l}{Face} & \multicolumn{1}{|c}{image (i)}\\
\hline
\multicolumn{1}{l}{OCR} & \multicolumn{1}{|c}{image (i)}\\
\hline
\multicolumn{1}{l}{RoBERTa} & \multicolumn{1}{|c}{text (t), OCR text (ocr)}\\
\hline
\multicolumn{1}{l}{Index Embedding (IE)} & \multicolumn{1}{|c}{object classes (obj), image naturalness (nat),}\\
\multicolumn{1}{l}{} & \multicolumn{1}{|c}{text concreteness (con)}\\
\Xhline{2\arrayrulewidth}
\end{tabular}%
}
\end{center}
\caption{Features and inputs types used for our experiments. Image naturalness, and text concreteness are metadata provided by Ads-Parallelity Dataset.}
\label{tab:feat}
\end{table}



We extract various features for both images and text using off-the-shelf models.
Table~\ref{tab:feat} summarizes all the input types for different feature types.

\textbf{WSL.} We use the output from the second-to-last FC layer of the ResNeXt-101 model ($32\times16d$, pretrained on 940 million public images and ImageNet1K dataset)~\cite{wslimageseccv2018} as what we call a \textit{whole image feature}. The dimension of this feature is 2048.

\textbf{Face.} 
We use MTCNN~\cite{zhang2016mtnnface} to detect faces from images, and for each detected face, we extract out that part of the image from the face bounding box and obtain per-detection face embeddings (dimension 256) from a pre-trained model (SE-ResNet-50-256D~\cite{Cao18vggface2}), trained on the VGG-Face2 dataset \cite{Cao18}.

\textbf{Optical Character Recognition (OCR).}
We use an existing OCR model from~\cite{borisyuk2018rosetta} to extract overlaid text from images.

\textbf{Objects (obj) and bounding boxes (bbox).}
We use Faster R-CNN with a ResNet-50 backbone with FPN~\cite{lin2017feature} trained on the COCO dataset~\cite{lin2014microsoft} with the 1x training schedule from detectron2 model zoos\footnote{~\url{https://github.com/facebookresearch/detectron2}}.
We extract the COCO categories with classification score threshold of $0.65$, and extract the WSL embeddings of the associated bounding boxes.

\textbf{Sentence embeddings (RoBERTa).}
To encode raw text, we employ \textbf{RoBERTa} (using the BERT-base architecture)~\cite{liu2019roberta} to extract sentence embeddings from text as well as any detected OCR text.
We use the first output (dimension 768) of the final layer from the model download from fairseq~\cite{ott2019fairseq}. We concatenate all sentences as one for an individual sample, clipping the maximum sentence length of tokens to 512.

\textbf{Index Embeddings (IE).}
For certain types of features, the representation is a simple class index; an integer in a finite range (e.g., an object class index).  To represent these as embeddings, we construct the encoder by borrowing from the text modeling domain.  When we encode text, we typically build a vocabulary of unique words and map each word to a corresponding integer index.  These integer indices are then mapped to dense feature embeddings by means of a learned lookup-table such that each index is a row in a $|V| \times D$ dense matrix, for a vocabulary of size $|V|$.  These are typically referred to as \textit{word embeddings} as a way to map raw text words to dense embeddings; in this way, we construct \textit{index embeddings} as a way to map discrete class indices to dense embeddings.  For situations where there are more than one indices for a sample, we use the model architecture in \cite{Kim14} to encode this sequence further into a single embedding.  For example, in the Ads-Parallelity Dataset, there are ``image naturalness'' and ``text concreteness'' discrete labels; these are similarly modeled as described here.

\subsection{Implementation}
We used Pytorch~\cite{paszke2017pytorch} to implement and train all the models on a single NVIDIA V100 GPU.
Adam optimization with decoupled weight decay~\cite{loshchilov2018decoupled} is used.
The learning rate is warmed up linearly from 0 to 0.001 during the first five epochs, and then decayed with a cosine annealing schedule over the remaining epochs.
Sigmoid cross-entropy loss is used for both the single- and multi-label classification tasks.
For imbalanced training datasets (MM-IMDb), we use class weights (inverse square root of class frequency \cite{wslimageseccv2018}) to balance out the loss.
To stabilize the training process, the bias for the last linear classification layer is initialized with $b = -\log\left(\left(1 - \pi\right)/\pi\right)$~\cite{focal_loss, cui2019classbalancedloss}, where the prior probability $\pi$ is set to 0.01. 

\textbf{Modality Encoders:} Each feature modality encoder $\phi$ is trained to encode to a common space with dimensions $D$ so that the various modalities can be pooled together.  For these experiments, we use $D_{Ads} = 32$ for the Ads-Parallelity dataset, and $D_{IMDb} = 1024$ for MM-IMDb. These are chosen empirically based on the amount of available data to train as well as complexity of the problem.  For each $\phi$ (e.g., corresponding to each feature modality), we use a single fully-connected layer with an ELU activation \cite{clevert15ELU}, using a dropout of $0.25$, each having a final dimensionality of $D$, given the dataset.  (Note: for some datasets, there are too many instances of a particular modality, for example, objects or faces.  In these cases, to keep computational constraints in mind, we sub-sample down to 10 bounding box detections when need be.)

\textbf{Deep Set Pooling:} We study various pooling schemes for $\psi$, including \textit{max}, \textit{sum}, and \textit{min}.  Though \textit{max} is the proposed choice for interpretable feature importance, in some cases other pooling schemes yield better results.  This represents a common trade-off between model interpretability and performance.  We discuss more about the pooling operators below.

\textbf{Set Predictors:} For each set encoding following $\psi$, we construct MLPs to represent $\rho$ for the final down-stream tasks.  The architectures for each are as follows: $\rho_{Ads}: [D_{Ads} \rightarrow 32 \rightarrow 2]$, and $\rho_{IMDb}: [D_{IMDb} \rightarrow 256 \rightarrow 128 \rightarrow 32]$.  In each case, the last output layer represents the total number of predicted classes for that dataset task.

\subsection{Results and Discussions}
\begin{table*}[t]
\scriptsize
\begin{center}
\begin{tabular}{ lcccccccc }
\Xhline{2\arrayrulewidth}
\multicolumn{1}{c}{ } &\multicolumn{1}{|c}{ }&\multicolumn{1}{|c}{ } &\multicolumn{1}{|c}{ } &\multicolumn{2}{|c}{\textbf{Overall}} &\multicolumn{1}{|c}{\textbf{Non-}}  & \multicolumn{2}{|c}{\textbf{Parallel}}\\
\cline{5-6}
\cline{8-9}
\multicolumn{1}{c}{\textbf{Method} } &\multicolumn{1}{|c}{\textbf{Modalities}} &\multicolumn{1}{|c}{\textbf{Features}} 
&\multicolumn{1}{|c}{\textbf{Pool Type}} & \multicolumn{1}{|c}{\textbf{Accuracy}} & \multicolumn{1}{c}{\textbf{AUC}} & \multicolumn{1}{|c}{\textbf{Parallel}} & \multicolumn{1}{|c}{\textbf{Equiv}} & \multicolumn{1}{c}{\textbf{Non-Equiv}} \\
\Xhline{2\arrayrulewidth}

\multicolumn{1}{c}{~\cite{zhang2018equal}} & \multicolumn{1}{|c}{img + txt} & \multicolumn{1}{|c}{-} &
\multicolumn{1}{|r}{-} & \multicolumn{1}{|r}{65.500} & \multicolumn{1}{|r}{-} & \multicolumn{1}{|r}{63.300} & \multicolumn{1}{|r}{70.200} & \multicolumn{1}{|r}{65.500} \\
\Xhline{2\arrayrulewidth}
\multicolumn{1}{c}{Ours~\one} & \multicolumn{1}{|c}{img} 
& \multicolumn{1}{|r}{WSL(i)} & \multicolumn{1}{|c}{-}
& \multicolumn{1}{|r}{57.726} & \multicolumn{1}{|r}{54.508} 
& \multicolumn{1}{|r}{47.803} & \multicolumn{1}{|r}{76.103} & \multicolumn{1}{|r}{59.273}\\

\multicolumn{1}{c}{Ours~\two} & \multicolumn{1}{|c}{img} 
& \multicolumn{1}{|r}{WSL(i), RoBERTa(ocr)} & \multicolumn{1}{|c}{$\psi_{max}$}
& \multicolumn{1}{|r}{61.721} & \multicolumn{1}{|r}{59.188} 
& \multicolumn{1}{|r}{46.553} & \multicolumn{1}{|r}{81.912} & \multicolumn{1}{|r}{71.283}\\

\multicolumn{1}{c}{Ours~\three} & \multicolumn{1}{|c}{img} 
& \multicolumn{1}{|r}{WSL(i), RoBERTa(ocr), Face(i)} & \multicolumn{1}{|c}{$\psi_{max}$}
& \multicolumn{1}{|r}{62.043} & \multicolumn{1}{|r}{57.681} 
& \multicolumn{1}{|r}{51.667} & \multicolumn{1}{|r}{75.956} & \multicolumn{1}{|r}{68.766}\\

\multicolumn{1}{c}{Ours~\four} & \multicolumn{1}{|c}{img} 
& \multicolumn{1}{|r}{WSL(i), RoBERTa(ocr), Face(i), CNN-text(objs)} & \multicolumn{1}{|c}{$\psi_{max}$}
& \multicolumn{1}{|r}{62.029} & \multicolumn{1}{|r}{58.648} 
& \multicolumn{1}{|r}{55.985} & \multicolumn{1}{|r}{76.912} & \multicolumn{1}{|r}{58.913}\\

\multicolumn{1}{c}{Ours~\five} & \multicolumn{1}{|c}{img} 
& \multicolumn{1}{|r}{WSL(i), RoBERTa(ocr), Face(i), CNN-text(objs), WSL(bbox)} & \multicolumn{1}{|c}{$\psi_{max}$}
& \multicolumn{1}{|r}{61.707} & \multicolumn{1}{|r}{59.402} 
& \multicolumn{1}{|r}{40.284} & \multicolumn{1}{|r}{84.412} & \multicolumn{1}{|r}{81.217}\\

\multicolumn{1}{c}{ } & \multicolumn{1}{|c}{ } 
& \multicolumn{1}{|r}{WSL(i), RoBERTa(ocr), Face(i),} & \multicolumn{1}{|c}{}
& \multicolumn{1}{|r}{ } & \multicolumn{1}{|r}{} 
& \multicolumn{1}{|r}{} & \multicolumn{1}{|r}{} & \multicolumn{1}{|r}{}\\
\multicolumn{1}{c}{Ours~\six} & \multicolumn{1}{|c}{img}
& \multicolumn{1}{|r}{IE(objs), WSL(bbox), IE(nat)}  & \multicolumn{1}{|c}{$\psi_{max}$}
& \multicolumn{1}{|r}{63.909} & \multicolumn{1}{|r}{61.300} 
& \multicolumn{1}{|r}{58.371} & \multicolumn{1}{|r}{72.426} & \multicolumn{1}{|r}{66.846}\\
\hline

\multicolumn{1}{c}{Ours~\seven} & \multicolumn{1}{|c}{txt}
& \multicolumn{1}{|r}{RoBERTa(t)} & \multicolumn{1}{|c}{-}
& \multicolumn{1}{|r}{46.777} & \multicolumn{1}{|r}{55.889} 
& \multicolumn{1}{|r}{\textbf{81.487}} & \multicolumn{1}{|r}{36.718} & \multicolumn{1}{|r}{28.294}\\

\multicolumn{1}{c}{Ours~\eight} & \multicolumn{1}{|c}{txt}
& \multicolumn{1}{|r}{RoBERTa(t), IE(con)} & \multicolumn{1}{|c}{$\psi_{max}$}
& \multicolumn{1}{|r}{72.167} & \multicolumn{1}{|r}{65.720} 
& \multicolumn{1}{|r}{67.140} & \multicolumn{1}{|r}{87.941} & \multicolumn{1}{|r}{65.776}\\
\hline

\multicolumn{1}{c}{ } & \multicolumn{1}{|c}{ }
& \multicolumn{1}{|r}{ }  & \multicolumn{1}{|c}{$\psi_{max}$}
& \multicolumn{1}{|r}{71.552} & \multicolumn{1}{|r}{73.326} 
& \multicolumn{1}{|r}{64.583} & \multicolumn{1}{|r}{\textbf{91.544}} & \multicolumn{1}{|r}{65.171}\\
\multicolumn{1}{c}{Ours~\nine} & \multicolumn{1}{|c}{img + txt}
& \multicolumn{1}{|r}{WSL(i), IE(nat), RoBERTa(ocr), RoBERTa(t), IE(con)}  & \multicolumn{1}{|c}{$\psi_{min}$}
& \multicolumn{1}{|r}{73.058} & \multicolumn{1}{|r}{70.113} 
& \multicolumn{1}{|r}{65.170} & \multicolumn{1}{|r}{90.294} & \multicolumn{1}{|r}{71.275}\\
\multicolumn{1}{c}{ } & \multicolumn{1}{|c}{ }
& \multicolumn{1}{|r}{ } & \multicolumn{1}{|c}{$\psi_{sum}$}
& \multicolumn{1}{|r}{\textbf{76.713}} & \multicolumn{1}{|r}{\textbf{77.788}} 
& \multicolumn{1}{|r}{75.152} & \multicolumn{1}{|r}{88.971} & \multicolumn{1}{|r}{67.451}\\
\hline

\multicolumn{1}{c}{ } & \multicolumn{1}{|c}{ }
& \multicolumn{1}{|r}{ }  & \multicolumn{1}{|c}{$\psi_{max}$}
& \multicolumn{1}{|r}{70.311} & \multicolumn{1}{|r}{67.464} 
& \multicolumn{1}{|r}{60.246} & \multicolumn{1}{|r}{86.691} & \multicolumn{1}{|r}{\textbf{73.840}}\\
\multicolumn{1}{c}{Ours~\ten} & \multicolumn{1}{|c}{img + txt}
& \multicolumn{1}{|r}{WSL(i), IE(nat), WSL(bbox),  RoBERTa(ocr),}  & \multicolumn{1}{|c}{$\psi_{min}$}
& \multicolumn{1}{|r}{71.561} & \multicolumn{1}{|r}{67.982} 
& \multicolumn{1}{|r}{63.409} & \multicolumn{1}{|r}{91.471} & \multicolumn{1}{|r}{67.590}\\
\multicolumn{1}{c}{ } & \multicolumn{1}{|c}{ } 
& \multicolumn{1}{|r}{Face(i), RoBERTa(t), IE(con)} & \multicolumn{1}{|c}{$\psi_{sum}$}
& \multicolumn{1}{|r}{73.716} & \multicolumn{1}{|r}{75.460} 
& \multicolumn{1}{|r}{69.545} & \multicolumn{1}{|r}{91.471} & \multicolumn{1}{|r}{63.709}\\
\hline

\multicolumn{1}{c}{ } & \multicolumn{1}{|c}{ }
& \multicolumn{1}{|r}{ } & \multicolumn{1}{|c}{$\psi_{max}$}
& \multicolumn{1}{|r}{71.845} & \multicolumn{1}{|r}{73.334} 
& \multicolumn{1}{|r}{63.371} & \multicolumn{1}{|r}{90.147} & \multicolumn{1}{|r}{70.229}\\
\multicolumn{1}{c}{Ours~\eleven} & \multicolumn{1}{|c}{img + txt} 
& \multicolumn{1}{|r}{WSL(i), IE(nat), objs, bbox, RoBERTa(ocr),} & \multicolumn{1}{|c}{$\psi_{min}$}
& \multicolumn{1}{|r}{71.262} & \multicolumn{1}{|r}{73.428}
& \multicolumn{1}{|r}{60.928} & \multicolumn{1}{|r}{90.221} & \multicolumn{1}{|r}{72.377}\\
\multicolumn{1}{c}{} & \multicolumn{1}{|c}{ }
& \multicolumn{1}{|r}{Face(i), transcription (r), IE(con)} & \multicolumn{1}{|c}{$\psi_{sum}$}
& \multicolumn{1}{|r}{74.923} & \multicolumn{1}{|r}{75.045} 
& \multicolumn{1}{|r}{72.064} & \multicolumn{1}{|r}{\textbf{91.544}} & \multicolumn{1}{|r}{64.118}\\
\Xhline{2\arrayrulewidth}

\end{tabular}
\end{center}
\caption{Ads-Parallelity results. Average accuracy and area under the ROC curve over entire datasets, and accuracies for three fine-grained classes (Non-Parallel: text and image convey different meanings; Parallel-Equiv: they are completely equivalent; Parallel-Non-Equiv: they express same ideas but in different levels of details.) are reported. The proposed Deep Multi-Modal Sets is able to outperform the baseline method by a large margin.}
\label{tab:ads_results}
\end{table*}

\begin{figure*}
    \centering
    \includegraphics[width=0.82\textwidth]{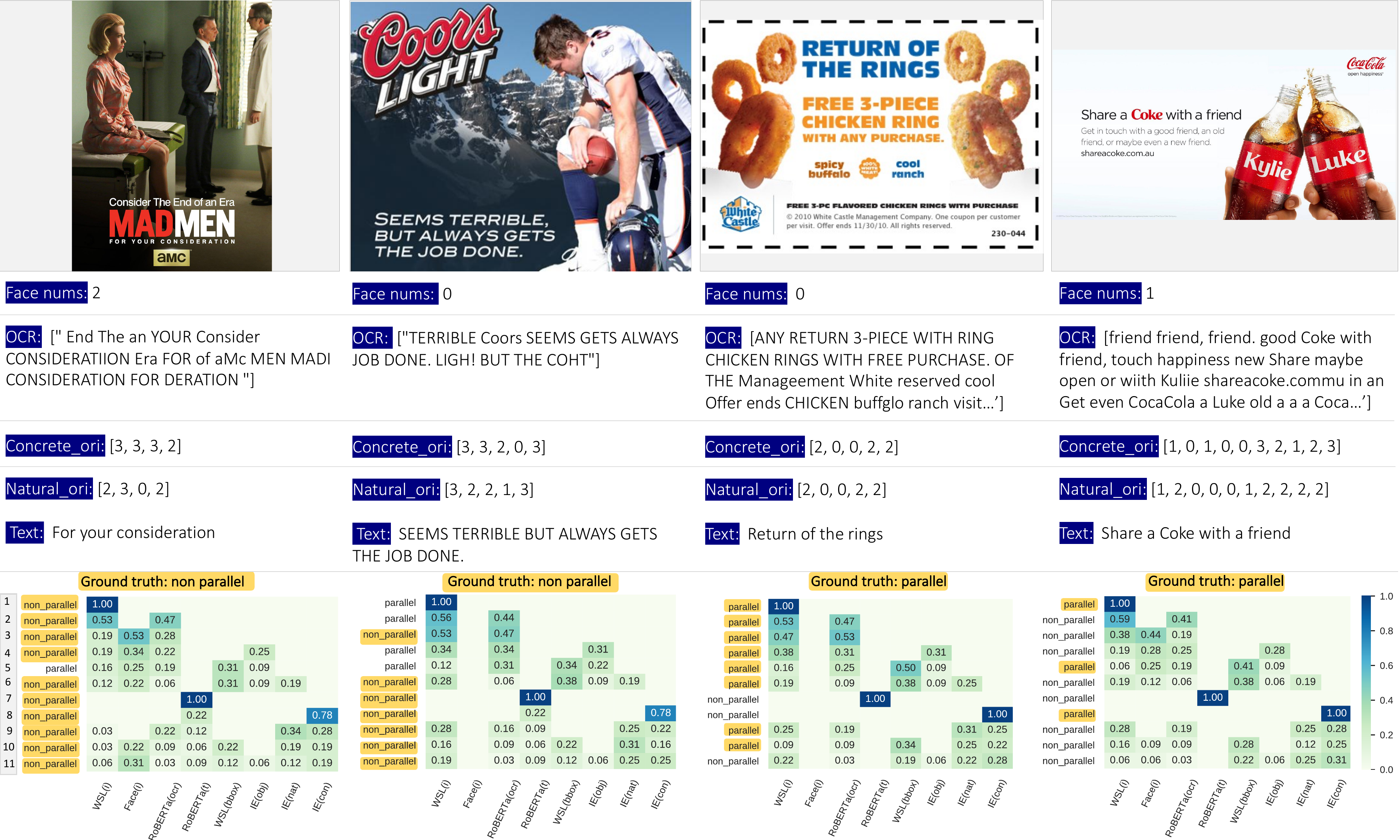}
    \caption{Selective examples from Ads-Parallelity. Input of this dataset include image, text, faces, OCR text, \etc. Last row of the figure presents the Feature Importance Matrices (FIM) for each model (referenced in Table~\ref{tab:ads_results}). Best view digitally.}
    \label{fig:adsresults}
\end{figure*}

\begin{table*}[t]
\scriptsize
\begin{center}
\begin{tabular}{ lcccccc }
\Xhline{2\arrayrulewidth}
\multicolumn{1}{c}{\textbf{Method}} &\multicolumn{1}{|c}{\textbf{Modalities}} &\multicolumn{1}{|c}{\textbf{Features}}
&\multicolumn{1}{|c}{\textbf{Pool Type}}
& \multicolumn{1}{|c}{\textbf{F1-Micro}} & \multicolumn{1}{|c}{\textbf{F1-Macro}} & \multicolumn{1}{|c}{\textbf{F1-Samples}} \\
\Xhline{2\arrayrulewidth}
\multicolumn{1}{c}{~\cite{Arevalo17}} & \multicolumn{1}{|c}{img + txt} &
\multicolumn{1}{|r}{-}
& \multicolumn{1}{|r}{-}
& \multicolumn{1}{|r}{0.6300} & \multicolumn{1}{|r}{0.5410} & \multicolumn{1}{|r}{0.6300}\\
\multicolumn{1}{c}{~\cite{kiela2018efficient}} & \multicolumn{1}{|c}{img + txt} &
\multicolumn{1}{|r}{-}
& \multicolumn{1}{|r}{-}
& \multicolumn{1}{|r}{0.6230} & \multicolumn{1}{|r}{-} & \multicolumn{1}{|r}{-}\\
\multicolumn{1}{c}{~\cite{prezra2019mfas}} & \multicolumn{1}{|c}{img + txt} &
\multicolumn{1}{|r}{-}
& \multicolumn{1}{|r}{-}
& \multicolumn{1}{|r}{-} & \multicolumn{1}{|r}{0.5568} & \multicolumn{1}{|r}{-}\\
\multicolumn{1}{c}{~\cite{kiela2019supervised}} & \multicolumn{1}{|c}{img + txt} &
\multicolumn{1}{|r}{-}
& \multicolumn{1}{|r}{-}
& \multicolumn{1}{|r}{0.6640} & \multicolumn{1}{|r}{0.6110} & \multicolumn{1}{|r}{-}\\
\Xhline{2\arrayrulewidth}

\multicolumn{1}{c}{Ours~\one} & \multicolumn{1}{|c}{img}
& \multicolumn{1}{|r}{WSL(i)}
& \multicolumn{1}{|r}{-}
& \multicolumn{1}{|r}{0.5253} & \multicolumn{1}{|r}{0.3791}& \multicolumn{1}{|r}{0.5227} \\

\multicolumn{1}{c}{Ours~\two} & \multicolumn{1}{|c}{img}
& \multicolumn{1}{|r}{WSL(i), Face(i), IE(obj)}
& \multicolumn{1}{|r}{$\psi_{max}$}
& \multicolumn{1}{|r}{0.4945} & \multicolumn{1}{|r}{0.3570}& \multicolumn{1}{|r}{0.4931} \\

\multicolumn{1}{c}{Ours~\three} & \multicolumn{1}{|c}{img}
& \multicolumn{1}{|r}{WSL(i), Face(i), IE(obj), WSL(bbox)}
& \multicolumn{1}{|r}{$\psi_{max}$}
& \multicolumn{1}{|r}{0.4939} & \multicolumn{1}{|r}{0.3566}& \multicolumn{1}{|r}{0.4933} \\

\multicolumn{1}{c}{Ours~\four} & \multicolumn{1}{|c}{img}
& \multicolumn{1}{|r}{WSL(i), Face(i), IE(obj), WSL(bbox), IE(ocr), RoBERTa(ocr)}
& \multicolumn{1}{|r}{$\psi_{max}$}
& \multicolumn{1}{|r}{0.5150} & \multicolumn{1}{|r}{0.3856}& \multicolumn{1}{|r}{0.5173} \\
\hline

\multicolumn{1}{c}{Ours~\five} & \multicolumn{1}{|c}{txt}
& \multicolumn{1}{|r}{RoBERTa(t)}
& \multicolumn{1}{|r}{$\psi_{max}$}
& \multicolumn{1}{|r}{0.6699} & \multicolumn{1}{|r}{0.6011} & \multicolumn{1}{|r}{0.6714}\\
\hline

\multicolumn{1}{c}{ } & \multicolumn{1}{|c}{ } 
& \multicolumn{1}{|r}{ }
& \multicolumn{1}{|r}{$\psi_{max}$}
& \multicolumn{1}{|r}{0.6623} & \multicolumn{1}{|r}{0.5961}& \multicolumn{1}{|r}{0.6637} \\
\multicolumn{1}{c}{Ours~\six} & \multicolumn{1}{|c}{img + txt} 
& \multicolumn{1}{|r}{WSL(i), RoBERTa(t)}
& \multicolumn{1}{|r}{$\psi_{min}$}
& \multicolumn{1}{|r}{0.6709} & \multicolumn{1}{|r}{0.5929}& \multicolumn{1}{|r}{0.6710} \\
\multicolumn{1}{c}{ } & \multicolumn{1}{|c}{} & \multicolumn{1}{|r}{ }
& \multicolumn{1}{|r}{$\psi_{sum}$}
& \multicolumn{1}{|r}{0.6716} & \multicolumn{1}{|r}{0.5885}& \multicolumn{1}{|r}{0.6721} \\
\hline

\multicolumn{1}{c}{ } & \multicolumn{1}{|c}{ } & \multicolumn{1}{|r}{ }
& \multicolumn{1}{|r}{$\psi_{max}$}
& \multicolumn{1}{|r}{\textbf{0.6773}} & \multicolumn{1}{|r}{\textbf{0.6133}}& \multicolumn{1}{|r}{\textbf{0.6763}} \\

\multicolumn{1}{c}{Ours~\seven} & \multicolumn{1}{|c}{img + txt} 
& \multicolumn{1}{|r}{WSL(i), RoBERTa(t), WSL(bbox)}
& \multicolumn{1}{|r}{$\psi_{min}$}
& \multicolumn{1}{|r}{0.6673} & \multicolumn{1}{|r}{0.5871}& \multicolumn{1}{|r}{0.6690} \\

\multicolumn{1}{c}{ } & \multicolumn{1}{|c}{ } & \multicolumn{1}{|r}{ }
& \multicolumn{1}{|r}{$\psi_{sum}$}
& \multicolumn{1}{|r}{0.6677} & \multicolumn{1}{|r}{0.5965}& \multicolumn{1}{|r}{0.6665} \\
\hline

\multicolumn{1}{c}{ } & \multicolumn{1}{|c}{ } & \multicolumn{1}{|r}{ }
& \multicolumn{1}{|r}{$\psi_{max}$}
& \multicolumn{1}{|r}{0.6474} & \multicolumn{1}{|r}{0.5671}& \multicolumn{1}{|r}{0.6502} \\

\multicolumn{1}{c}{Ours~\eight} & \multicolumn{1}{|c}{img + txt} 
& \multicolumn{1}{|r}{WSL(i), RoBERTa(t), WSL(bbox), Face(i)}
& \multicolumn{1}{|r}{$\psi_{min}$}
& \multicolumn{1}{|r}{0.6302} & \multicolumn{1}{|r}{0.5532}& \multicolumn{1}{|r}{0.6310} \\

\multicolumn{1}{c}{ } & \multicolumn{1}{|c}{ } & \multicolumn{1}{|r}{ }
& \multicolumn{1}{|r}{$\psi_{sum}$}
& \multicolumn{1}{|r}{0.6664} & \multicolumn{1}{|r}{0.5948}& \multicolumn{1}{|r}{0.6665} \\
\hline

\multicolumn{1}{c}{ } & \multicolumn{1}{|c}{ } & \multicolumn{1}{|r}{ }
& \multicolumn{1}{|r}{$\psi_{max}$}
& \multicolumn{1}{|r}{0.6345} & \multicolumn{1}{|r}{0.5459}& \multicolumn{1}{|r}{0.6328} \\
\multicolumn{1}{c}{Ours~\nine} & \multicolumn{1}{|c}{img + txt } 
& \multicolumn{1}{|r}{WSL(i), RoBERTa(t), WSL(bbox), Face(i), IE(obj)}
& \multicolumn{1}{|r}{$\psi_{min}$}
& \multicolumn{1}{|r}{0.6544} & \multicolumn{1}{|r}{0.5688}& \multicolumn{1}{|r}{0.6526} \\
\multicolumn{1}{c}{ } & \multicolumn{1}{|c}{ } & \multicolumn{1}{|r}{ }
& \multicolumn{1}{|r}{$\psi_{sum}$}
& \multicolumn{1}{|r}{0.6688} & \multicolumn{1}{|r}{0.5884}& \multicolumn{1}{|r}{0.6666} \\
\hline

\multicolumn{1}{c}{ } & \multicolumn{1}{|c}{ } & \multicolumn{1}{|r}{ }
& \multicolumn{1}{|r}{$\psi_{max}$}
& \multicolumn{1}{|r}{0.6353} & \multicolumn{1}{|r}{0.5621}& \multicolumn{1}{|r}{0.6363} \\

\multicolumn{1}{c}{Ours~\ten} & \multicolumn{1}{|c}{img + txt } 
& \multicolumn{1}{|r}{WSL(i), RoBERTa(t), WSL(bbox), Face(i), IE(obj), RoBERTa(ocr)}
& \multicolumn{1}{|r}{$\psi_{min}$}
& \multicolumn{1}{|r}{0.6410} & \multicolumn{1}{|r}{0.5530}& \multicolumn{1}{|r}{0.6430} \\

\multicolumn{1}{c}{ } & \multicolumn{1}{|c}{ } & \multicolumn{1}{|r}{ }
& \multicolumn{1}{|r}{$\psi_{sum}$}
& \multicolumn{1}{|r}{0.6750} & \multicolumn{1}{|r}{0.6044}& \multicolumn{1}{|r}{0.6760} \\
\Xhline{2\arrayrulewidth}

\end{tabular}
\end{center}
\caption{MM-IMDb results. Micro F1, Macro F1, Samples F1 scores are reported. The proposed Deep Multi-Modal Sets is able to outperform the other methods.}
\label{tab:imdb_results}
\end{table*}

\begin{figure*}[!t]
    \centering
    \includegraphics[width=0.82\textwidth]{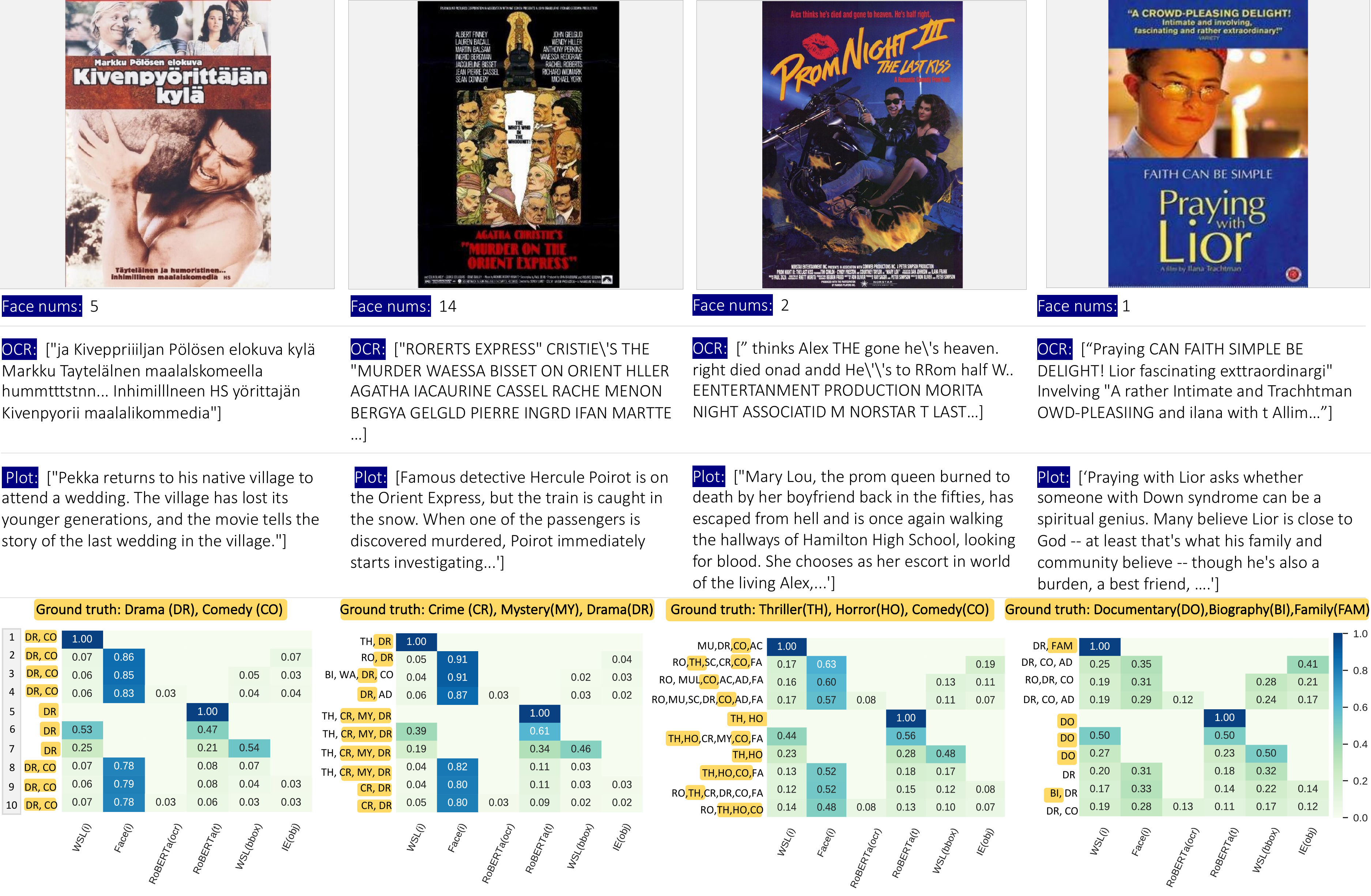}
    \caption{Selective examples from MM-IMDb. We list input for each sample including poster, faces, OCR text, and plot. Plots and OCR are abbreviated for visual effect. Last row of the figure presents the Feature Importance Matrices (FIM) for each model (referenced in Table~\ref{tab:imdb_results}). Predicted labels at y-axis are abbreviated as: Action (AC),  Adventure (AD), Biography (BI), Comedy (CO), Crime (CR), Documentary (DO), Drama (DR), Family (FAM), Fantasy (FA), Horror (HO), Music (MU), Musical (MUL), Mystery (MY), Romance (RO), Sci-Fi (SC), Thriller (TH), War (WA). Best view digitally.}
    \label{fig:imdbresults}
\end{figure*}

\textbf{Ads-Parallelity:} Table \ref{tab:ads_results} shows our results using Deep Multi-Modal Sets on the Ads-Parallelity dataset.  Here, we show ablation studies using various combinations of features and pooling schemes to best understand how different modalities interact with one another for this particular task and compare each to the current state-of-the-art shown in the top row \cite{zhang2018equal}.  The metrics used here are overall accuracy and AUC as well as individually for both the non-parallel and parallel tasks.  The best metrics are shown in bold, showing a noticeable increase in performance over the current SOTA.  

We find that, in general, and perhaps unsurprisingly, more features working together do help boost performance over uni-modal approaches.  This is an advantage to the proposed method.  It allows, in a very flexible manner, to support just about any type of feature as well as any number of instances of each modality, without increasing the model complexity.  In this way, the model can leverage more types of modalities jointly.  We stress that, though many features, such as WSL or OCR, typically have one occurrence in a given sample, modalities such as object detections can occur any number of times in an image, and the model need not know this beforehand; each is simply an instance in the set and there may be as few or many instances as naturally occur, without padding or place-holders.  This requires less up-front manual feature engineering.  In the end, our boost in performance ranges from \textbf{+11.213\%} on overall accuracy to \textbf{+21.344\%} on the parallel equivalence task.

We show some sample results from this experiment in Fig. \ref{fig:adsresults}.  Here, the top row shows sample images from the dataset.  Below that, we show the various features either extracted (e.g., number of faces, OCR overlaid text) or given (e.g., concreteness and naturalness labels, supplemental raw text input).  In the last row, we show the feature importance (FI) for each modality for that individual test sample after prediction, as it is considered in various feature combinations.  On the x-axis of the FI matrices (FIM), we label the individual feature modalities that are considered.  On the y-axis, we show the predicted label for each combination of features (shown per-row).  The top of each FIM shows the ground-truth for that given test sample.  

In the first column, we show an example "Mad Men" advertisement along with the supplemented text "For your consideration".  The ground truth label is given as \textit{non-parallel}, meaning that the image and text are not conveying the same message.  When looking at the FIM, all but one feature combination gets the prediction correct.  On further inspection, it seems that faces are quite important when included, as well as WSL(bbox) and naturalness and concreteness index labels.  This seems reasonable as faces are the major feature of the image, and comparing that to text would be a natural way to determine if they are parallel or not.  For the next column over (Coors Light ad), face importance goes down (empty means zero importance), which makes sense since the actor's face is present, but barely visible.  Here, WSL(i) as well as naturalness and concreteness dominate importance.  In the "White Castle" ad, WSL(bbox) (e.g., objects) as well as WSL(i) work well together with naturalness and concreteness, which seems reasonable given the content of the ad.  Finally, the rightmost column shows a mistake: most predictions get the label wrong.  When we inspect this, we see faces were incorrectly detected and the Roberta(t) is given zero importance too often.  This shows what happens when bad features occur, \textbf{however the FI gives us insight into why it is happening}. 

\textbf{MM-IMDb:} In Table \ref{tab:imdb_results}, we show results on MM-IMDb.  For this, we compare against 4 existing SOTA methods in \cite{Arevalo17, kiela2018efficient, prezra2019mfas, kiela2019supervised}.  To evaluate, as in those works, we use the F1-Micro, F1-Macro, and F1-Samples metrics.  Here, we can observe some slight trade-offs amongst the different input modalities when comparing image to text features.  We will note that the combination of images and text modalities outperforms either on their own, which highlights the need for multi-modal models whenever possible.  We are able to show improvement over all 4 prior methods on this dataset, with the best performing model using WSL(i), RoBERTa(t) and WSL(bbox) from objects.

We similarly show example predictions from this dataset in Fig. \ref{fig:imdbresults}, as we did above.  The left 3 columns show examples of correct predictions, where several of the feature combinations get most (or all) of the labels from this multi-label problem correct.  Faces are very helpful for the middle-left example, where clearly there are many faces present in the image.  For the middle-right example (Prom Night III), faces are also important as well as WSL(i) along with the text/plot description encoding.  The right-most column shows an example where the model performs poorly.  Here, no combination of features seems able to get all labels correct, though some cases are able to pick up on one of them.  The labels "Family" and "Biography" do not seem indicated in any of the visual features, and the signal from the raw text was not able to pick up for those mistakes enough to get all 3 labels correct.  Again, the analysis provided by this interpretable FI allows us to reason about model performance in a natural way.

\textbf{Pooling:} For Ads-Parallelity, it seems sum pooling slightly out-performs the other pooling schemes for overall accuracy and AUC, however in the individual tasks max-pooling occasionally edges out the others.  The difference is typically marginal, however this does showcase how different pooling schemes should be considered when designing the model architecture.  A single pooling method does not always out-perform the others in all cases.

In the case of MM-IMDb, max pooling is yielding the top results, interestingly with the same feature combination for all three metrics.  Again, the differences between pooling schemes is not large, but this study is still educational towards understanding pooling schemes.  Because the encoders are trained end-to-end with a particular pooling scheme in mind, they will attempt to optimize performance given the pooling operator.

\section{Conclusions}

In this paper we introduced a new model architecture for reasoning about multiple modalities that is more natural and less restrictive than previous concatenation-based fusion approaches.  Our method utilizes unordered sets in which any number of features are pooled together for down-stream tasks.  When features aren't present, we do not need any unnatural placeholders and when features occur variable number of times, we don't need padding.  We demonstrate new SOTA performance on challenging datasets and offer an interpretable method of feature importance when using max-pooling as the fusion scheme.  This allows us to reason about both correct and incorrect predictions at inference time.  Future work will include extension to video, as this is a common use of multi-modal modeling.

{\small
\bibliographystyle{ieee_fullname}
\bibliography{egbib}
}

\end{document}